# Flexible Computing Services for Comparisons and Analyses of Classical Chinese Poetry

[†]Chao-Lin Liu
Fairbank Center for Chinese Studies, Harvard University, USA
Department of Computer Science, National Chengchi University, Taiwan
chaolin@nccu.edu.tw

## Introduction

As for many civilizations, poetry is an essential part of Chinese literature. Poetry has influenced the development of the literature and language of both classical and vernacular Chinese. Certain of the words that we use today can be tracked all the way back to the Shijing (詩經/shi1 jing1/ [1]), c. 1046BC. Research on Chinese poetry is thus instrumental for understanding Chinese culture, and a lot of invaluable results have been accumulated over the past thousands of years from the study and analysis of Chinese poetry.

The availability of digital tools and resources enable researchers to compare and analyze the poetry from certain perspectives that were hard to achieve in the past. In many cases, we can verify the claims of previous researches with solid data, and, in others, we may enrich our understanding of the poetry.

The accessibility of increasingly larger datasets strengthens our research potential. In earlier stages of digital humanities, pioneers focused their work on Tang and Song poetry [2]. Now, we can access digitized texts of poems that were published in the periods from 1046BC to modern days.

Software tools allow us to study the data from a wide variety of perspectives in an efficient way. Search engines and information retrieval techniques [3] help us extract relevant texts from a large dataset. Then, researchers can employ domain knowledge for advanced studies with the use of additional tools.

In this paper, we showcase research results that we achieved by handling the available data with existing tools in flexible ways. We collected nine representative corpora of Chinese poetry, one each for a major dynasty in Chinese history between 1046BC and 1644AD. We list the corpora in Table 1, where we assign an acronym to each corpus for ease of reference [4]. We also show their Chinese names (**Collection**) and the periods of publication (**Time**). A collection for the Qing dynasty is unavailable yet because an editorial committee is still working toward the completion of this very challenging goal [5]. Excluding the punctuation marks that were added by the data providers, we have more than 16.5 million characters [6] in the corpora.

Table 1. The corpora of poetry of 1046BC-1644AD used in this study

| Acronym | Collection | Time | Acronym | Collection | Time |
|---|---|---|---|---|---|
| SJ | 詩經 | 1046-476BC | CV | 楚辭 | 475-221BC |
| HF | 漢賦(文選) | 202BC-420AD | PT | 先秦漢魏晉南北朝詩 | Before 589AD |
| CTP | 全唐詩 | 618-907AD | CSP | 全宋詩 | 960-1279AD |
| CSL | 全宋詞 | 960-1279AD | YSX | 元詩選 | 1271-1368AD |
| LCSJ | 列朝詩集 | 1368-1644AD | | | |

Table 2. The frequencies of selected words used in poems of LSY, LB, DM, and MF

|   | LSY | LB | DM | DF |   | LSY | LB | DM | DF |
|---|---|---|---|---|---|---|---|---|---|
| 春風;秋風 | 14;2;16 | 72;26;98 | 18;11;29 | 19;30;49 | 春草;秋草 | 0;0;0 | 15;12;27 | 2;1;3 | 13; 5;18 |
| 春水;秋水 | 2; 3; 5 | 3;10;13 | 4; 5; 9 | 8;12;20 | 春色;秋色 | 0;1;1 | 9;11;20 | 3;6;9 | 20; 7;27 |
| 春月;秋月 | 0; 0; 0 | 0;40;40 | 0; 0; 0 | 0; 4; 4 | 春來;秋來 | 4;1;5 | 0; 3; 3 | 2;6;8 | 8; 6;14 |
| 春日;秋日 | 2; 2; 4 | 2; 1; 3 | 3; 1; 4 | 13; 5;18 | 春光;秋光 | 2;1;3 | 6; 0; 6 | 2;3;5 | 9; 1;10 |
| 春山;秋山 | 2; 0; 2 | 2; 6; 8 | 0; 4; 4 | 2; 5; 7 | 春天;秋天 | 1;0;1 | 2; 2; 4 | 0;0;0 | 5;11;16 |
| 春雨;秋雨 | 0; 2; 2 | 0; 2; 2 | 3; 3; 6 | 4; 4; 8 | 春江;秋江 | 0;1;1 | 1; 2; 3 | 0;2;2 | 6; 2; 8 |

By flexibly integrating and migrating tools to offer new functions, we can provide researchers with opportunities to investigate Chinese poetry from new perspectives. In the first example, we show a new way to compare the ways that poets use words in their poems. In the second, with our own tools, we can find shared collocations and patterns of poems in different corpora, and this capability allows us to study and compare the styles of individual poets and their dynasties.

## A Multi-Faceted Comparison

Jiang [7] compared the usage of "wind" (風/feng1/) and "moon" (月/yue4/) in the poems of two of the most famous poets, Li Bai (李白/li3 bai2/) and Du Fu (杜甫/du4 fu4/), of the Tang Dynasty [8] by comparing the contents of selected poems. Liu et al. [9] listed the frequencies of frequent words that used "wind" and "moon" in Li's and Du's poems. The numerical comparison shows the differences of the poets in a vivid way.

The software tools can be designed so that we can inspect not just the original poems or the raw statistics about the original poems, but also more complex comparisons.

Table 2 lists the frequencies of frequent bigrams [6] that appeared in the poems of four poets, i.e., LSY (for 李商隱/li3 shang1 yin3/), LB (for Li Bai), DM (for 杜牧/du4 mu4/), and DF (for Du Fu)[10]. These bigrams are special in that they are formed by concatenating either "春"/chun1/ or "秋"/qiu1/ [11] with another character, and they represent something related to "spring" and "autumn", respectively [12]. The numbers "14;2;16" in the row of "春風;秋風" and in the column for "LSY" indicate that we have 14 and 2 of LSY's poems in which "春風" and "秋風" were used, respectively. "16" is the sum of 14 and 2.

The statistics in Table 2 shed light on the differences in word preferences among the poets. Note that the samples in Table 2 are limited, and that a close reading is necessary to reach any further interpretations. Despite these limitations, we still can explore comparisons from various perspectives. "春風" and "秋風" are the most common choices among all of the rows [13]. In contrast, "春月" and "秋月" were not as popular [14], and none of the poets used "春月". In terms of personal preference, "春風" appeared in LB's poems three times often than "秋風". The results of LSY are similar to those fore LB, but DF seems to prefer "秋風" instead[15].

The entries that have "0"s can be linked to strong personal preferences. For instance, LB did not use "春雨" and "春來", while he did use "秋雨" and "秋來". DM is special in that he did not use "春天" or "秋天".

We can provide different ways to compare the styles of poets, e.g., converting the frequencies in Table 2 to probabilities of seeing the same word in the poems. By building a vector space representation [3] for each poet, we can calculate a score of similarity for style as in many other researches.

## Networking Names and Words

In addition to comparing the words of the famous poets, we may also attempt to compare the words and themes of the poems that were produced by friends. We can look up whether two poets were friends in professional databases like the China Biographical Database (CBDB) [16]. A database like the CBDB can also provide alternative names of poets so that we may algorithmically find friendships among poets by checking their writings [9]. After identifying a group of poets who were friends, we can investigate whether the words, styles, and themes of their poems are related. A procedure such as which we used to produce statistics like those in Table 2 may be useful.

A poet may be influenced by another poet even though they were not personally acquainted. It is believed that poets of the same school of poetry [17] share similar styles or words. Hence, information about the membership of a school of poetry provides a starting point for an investigation.

We may also search for poets who shared the same words and collocations in their poems as a clue for an indirect friendship. Given the millions of characters in our corpora, we need to have an efficient mechanism to identify poems that shared collocations and patterns [18], and using our own tools, we can precisely identify words that were shared by poems of different poets and of different dynasties [19].

The ability to identify the shared words between individual poets also automatically allows us to compare the patterns that are frequently shared between any two corpora. In Figure 1, two words are connected if they frequently co-occurred in poems. Part (a) shows the shared collocations in poems in the YSX and CTP, and (b) is for the shared collocations in poems of the LCSJ and CTP. The differences between (a) and (b) indicate that the highly shared collocations changed from dynasty to dynasty, i.e., from Tang to Yuan and from Tang to Ming. A collocation with a different word may suggest that the word contributes a different sense in the poems, e.g., "春風-桃花" and "春風-何處" in (b), and this can be verified by reading the poems that used these collocations. Sometimes, the links suggest replaceable words, for instance, both "千里" and "萬里" can go with "十年" in both (a) and (b). It should be noted that the collocations often carry information about the imagery of the poems.

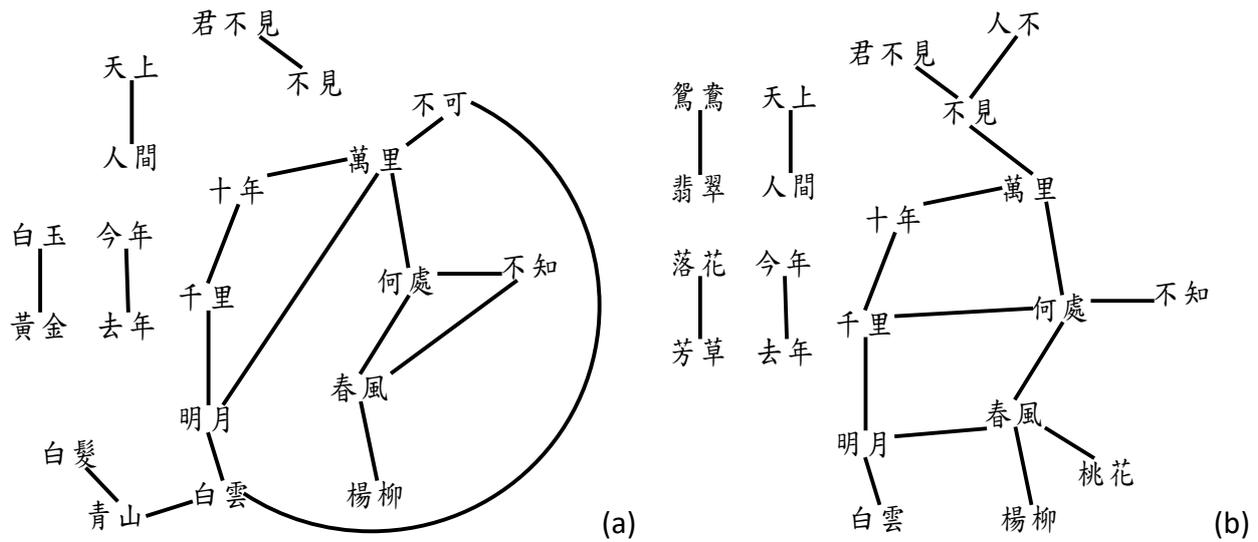

Figure 1. Frequently shared collocations between poems of two corpora (a) YSX and CTP (b) LCSJ and CTP

## Concluding Remarks

We briefly discussed how we studied two new research problems by flexible applications of our tools. The new tools provide new forms of data as in Table 2 and Figure 1 for interesting and useful research. We are working toward an in-depth understanding of Chinese words by studying when, who [18], and how the words [20] and their collocations and patterns were used in Chinese poetry, and our tools will help domain experts study challenging and interesting problems about it [21]. We also hope that the information and visualization that we have found and established for words can contribute to an interactive version of the complete Chinese lexicon [22].

## Acknowledgments

This research was supported in part by the grant 104-2221-E-004-005-MY3 from the Ministry of Science and Technology of Taiwan, the grant USA-HAR-105-V02 from the Top University Strategic Alliance of Taiwan, and the Senior Fulbright Research Grant 2016-2017.